\documentclass[preprint,12pt]{elsarticle}

\usepackage{amssymb}
\usepackage{array}
\usepackage{subfigure}
\usepackage{subcaption}
\usepackage{graphicx, wrapfig}
\usepackage[font=scriptsize,skip=0pt]{caption}
\usepackage{comment}
\usepackage{hyperref}
\usepackage{amsmath}
\usepackage{bm}
\usepackage{xcolor}
\usepackage[numbers]{natbib}
\usepackage{tabularx}
\usepackage{algorithm}
\usepackage{algpseudocode}
\usepackage{booktabs} 
\usepackage{array} 

\begin{document}

\begin{frontmatter}

\title{Enhanced accuracy through ensembling of randomly initialized auto-regressive models for time-dependent PDEs}

\affiliation[a]{organization={Department of Civil and Systems Engineering, Johns Hopkins University},
            addressline={3400 N. Charles Street}, 
            city={Baltimore},
            postcode={21218}, 
            state={MD},
            country={USA}}

\author{Ishan Khurjekar$^{a}$}
\ead{ishan.khurjekar@jhu.edu}

\author{Indrashish Saha$^{a}$}
\ead{isaha3@jhu.edu}

\author{Lori Graham-Brady$^{a}$}
\ead{lori@jhu.edu}

\author{Somdatta Goswami$^{a}$\corref{cor1}}
\ead{somdatta@jhu.edu}
\cortext[cor1]{Corresponding author}

\maketitle

\begin{abstract} 
Systems governed by partial differential equations (PDEs) require computationally intensive numerical solvers to predict spatiotemporal field evolution. While machine learning (ML) surrogates offer faster solutions, autoregressive inference with ML models suffer from error accumulation over successive predictions, limiting their long-term accuracy. We propose a deep ensemble framework to address this challenge, where multiple ML surrogate models with random weight initializations are trained in parallel and aggregated during inference. This approach leverages the diversity of model predictions to mitigate error propagation while retaining the autoregressive strategies ability to capture the system's time dependent relations. We validate the framework on three PDE-driven dynamical systems - stress evolution in heterogeneous microstructures, Gray-Scott reaction-diffusion, and planetary-scale shallow water system - demonstrating consistent reduction in error accumulation over time compared to individual models. Critically, the method requires only a few time steps as input, enabling full trajectory predictions with inference times significantly faster than numerical solvers. Our results highlight the robustness of ensemble methods in diverse physical systems and their potential as efficient and accurate alternatives to traditional solvers. The codes for this work are available on GitHub (\href{https://github.com/Graham-Brady-Research-Group/AutoregressiveEnsemble_SpatioTemporal_Evolution}{Link}).
\end{abstract}

\begin{graphicalabstract}
\centering
\includegraphics[width=0.65\textwidth]{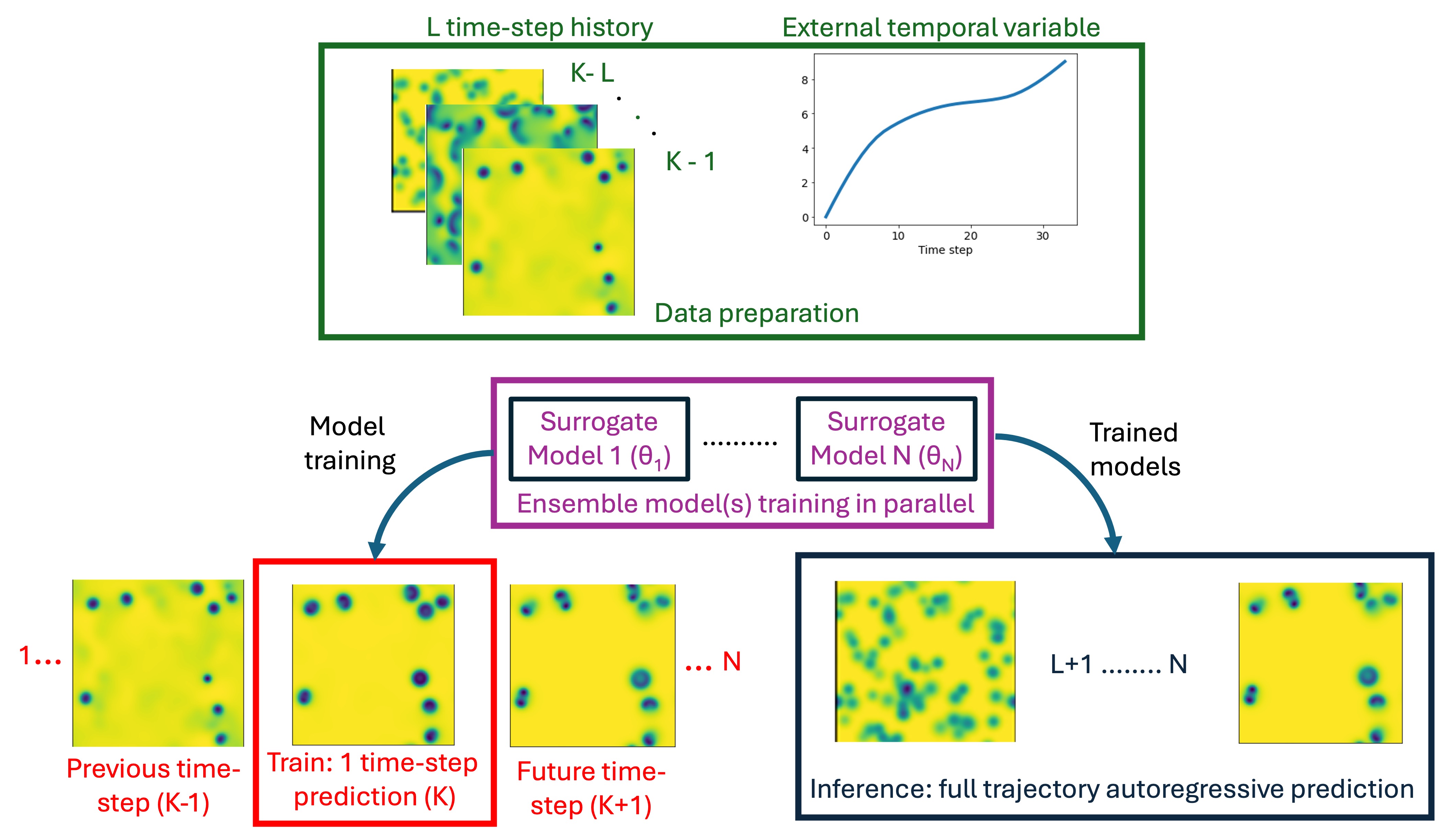}
    \label{fig:enter-label}
\end{graphicalabstract}

\begin{highlights}
\item An ensemble of machine learning surrogate models is initialized with random weights and trained in parallel for spatiotemporal field evolution tasks.
\item The ensemble approach leads to reduced error accumulation for the field evolution task in different PDE-driven dynamic systems.
\end{highlights}


\end{frontmatter}

\section{Introduction}
\label{sec:intro}

Accurate simulation of complex physical systems often relies on solving time-dependent partial differential equations (PDEs) using traditional numerical methods (finite difference, finite element, etc.), but these approaches can be prohibitively expensive when high resolution or many repeated queries are required. In many engineering and scientific applications (e.g. design optimization, uncertainty quantification), it is therefore desirable to build fast data-driven surrogate models that approximate the PDE solution operator. Deep neural networks have emerged as a powerful class of surrogates, capable of learning high-dimensional mappings from current system states to future states with far less online computation than conventional solvers. These ML-based simulators can leverage large training datasets to encode complex dynamics and thus provide flexible, real-time prediction of spatiotemporal evolution.

For time-dependent problems, a common strategy is to train an autoregressive neural model that predicts the next time step (or increment) given the current state, and then to apply it recursively to generate a full time trajectory. This one-step training approach is often more accurate and grounded in the physics of the system than trying to predict far-future states directly. However, a well-known challenge is that small one-step errors accumulate over repeated predictions. In other words, the iterative use of a learned time update operator tends to compound its own prediction errors, degrading stability and accuracy when the model is rolled out for long horizons. 

To address this challenge, several architectural innovations have been proposed. Recurrent networks (LSTM, GRU), attention-based transformers, and memory-augmented models have been introduced to improve temporal coherence. In the context of PDEs, neural operator (NO) frameworks such as DeepONet~\cite{lu2021learning} and Fourier Neural Operator~\cite{lifourier} offer a means to preserve spatial structures while modeling temporal evolution. Yet, NOs too have shown vulnerability to error accumulation in autoregressive settings~\cite{goswami2024learning}. Consequently, recent work has sought to enhance these models via architectural modifications \cite{diab2025temporal, robinson2022physics}, hybrid schemes with numerical integrators \cite{nayak2025ti}, and RNN-based sequence learners on top of NO outputs \cite{michalowska2024neural,he2024sequential}. While effective, these solutions often introduce architectural complexity and increase tuning burden.

In this work, we propose a conceptually simple yet effective alternative: mitigating long-term prediction error through deep ensembling. Rather than modifying the architecture, we train multiple instances of the same autoregressive model, each with a different random initialization. Due to the stochastic nature of training (e.g., with stochastic gradient descent), these models converge to different local minima and yield diverse prediction behaviors \cite{sutskever2013importance, huang2020dynamics}. By averaging the predictions from this ensemble at each time step, we reduce individual model biases and suppress cumulative error. This approach is grounded in classical ensemble learning theory, which shows that aggregating diverse learners improves robustness and generalization \cite{breiman2001random}. While ensemble methods such as bagging, boosting, and Mixture of Experts (MoE) often rely on explicit diversity through data splits or gating mechanisms \cite{jacobs1991adaptive}, we show that random initialization alone can induce sufficient diversity to enhance performance. Deep ensembles have been shown to improve accuracy, uncertainty calibration, and out-of-distribution robustness \cite{lakshminarayanan2017simple, zou2025learning, khurjekar2022sim, tripura2023foundational}; we adapt this principle for the specific challenge of long-horizon PDE forecasting.

We evaluate our ensemble strategy on three representative PDE systems: (i) stress field evolution in a two-phase microstructure under increasing strain (solid mechanics), (ii) the shallow water equations (fluid dynamics), and (iii) the Gray-Scott reaction–diffusion (chemical kinetics). These problems span a range of dynamical behaviors - from smooth deformations to wave propagation and pattern formation. In all cases, our ensemble-based surrogates consistently outperform single-model rollouts in long-horizon accuracy, without requiring any architectural modification, memory augmentation, or additional training objectives.

\section{Methods} \label{sec:methods}

Consider a 2D physical field $\mathbf{p}(x,y,t)$ defined over a continuous spatiotemporal domain, subjected to an external time-dependent forcing function $e(t)$. Numerical solutions typically assume a discretization of $\mathbf{p}(x,y,t)$ as $\mathbf{p}(\bf{x},\bf{y},\bf{t})$, where the components of vectors $\mathbf{x}$, $\mathbf{y}$, and $\mathbf{t}$ are $x_i: i\in{1,2,\ldots,N_x}$, $y_j:j\in{1,2,\ldots,N_y}$, and $t_k:k\in{1,\ldots,N_T}$, $N_{x} \times N_{y}$ are the spatial grid dimensions and $N_T$ is the number of time steps. Assuming uniform discretization $x_i=i\Delta x$, $y_j=j\Delta y$, $t_k=k\Delta t$, where $\Delta x$, $\Delta y$, and $\Delta t$ are the spatial and temporal resolutions, respectively. The forcing function $e(t)$ is similarly discretized as $e(\mathbf{t})$, or $e(t_k)$. In particular, numerical solvers attempt to find the solution at the next future state $\mathbf{p}(\mathbf{x}, \mathbf{y},t_{k+1})$ based on the solution history $\mathbf{p}(\mathbf{x},\mathbf{y},t_{k}): k\in{1,\ldots N_{T}-1}$ and the forcing function $e(t_{k+1}): k\in{1,\ldots N_{T}-1}$. Representing the full-time history of both the solution and the forcing function is typically infeasible; therefore, the solution is simplified through the use of state variables that provide a reduced-order representation of the forcing function and solution history. Such an approach allows the solution $\mathbf{p}(\mathbf{x},\mathbf{y},t_{k+1})$ to be found based only on the solution at the previous time step $t_k$, the current values of the state variables, and the current value of the forcing function $e(t_{k+1})$. Even with this simplified representation, calculation of the solution and the state variables at each time step is typically expensive, for example, requiring solutions involving high-dimensional matrices and iterative solvers. 

More recently, ML representations of the solution have proven to be much more efficient, with accuracy approaching that of the physics-based numerical solvers. Researchers have proposed sequence-to-sequence mapping approaches using transformer \cite{li2023scalable,hemmasian2024multi,wu2024transolver, geneva2022transformers}, CNNs \cite{saha2024prediction}, or Long Short Term Memory (LSTM) \cite{michalowska2023don} architectures to predict full trajectory rollouts. But such an approach assumes that the external conditions for the full simulation time are known in advance, limiting the real-world applicability. A more realistic approach is to predict the solution field for one time step at a time, using the historical information from previous time steps (autoregressive).

Specifically, consider an autoregressive ML surrogate model trained for the temporal evolution of the solution field. Mathematically, this is expressed as:
\begin{equation}\label{eq:surrogate_model}
 \mathbf{p}(\mathbf{x},\mathbf{y},t_{k+1}) = \mathcal{M} \{\mathbf{p}(\mathbf{x},\mathbf{y},t_k) ~, ~ e(t_{k+1}) \},
\end{equation}
where $\mathcal{M}$ denotes the ML-model. To avoid the need to calculate and track state variables, the accuracy of this autoregressive model is improved by considering not just one previous time step, but the previous $L$ time step history, assuming that this subset of time steps contains a sufficient amount of the information necessary to make an accurate prediction for the next time step \cite{oommen2022learning,cao2024deep}. This transforms Eq.~\ref{eq:surrogate_model} into,
\begin{equation}\label{eq:surrogate_model_history}
 \mathbf{p}(\mathbf{x},\mathbf{y},t_{k+1}) = \mathcal{M} \{\mathbf{p}(\mathbf{x},\mathbf{y},t_{k-L:k}) ~, ~ e(t_{k-L:k+1}) \},
\end{equation}
where $\mathbf{p}(\mathbf{x},\mathbf{y},t_{k-L:k})$ are the 2D maps of the physical field in the previous $L$ time steps and $e(t_{k-L:k+1})$ is the subset of the forcing function values in the $L$ previous time steps and the current time step.

\begin{figure}[!h]
    \centering
    \includegraphics[width=0.99\linewidth]{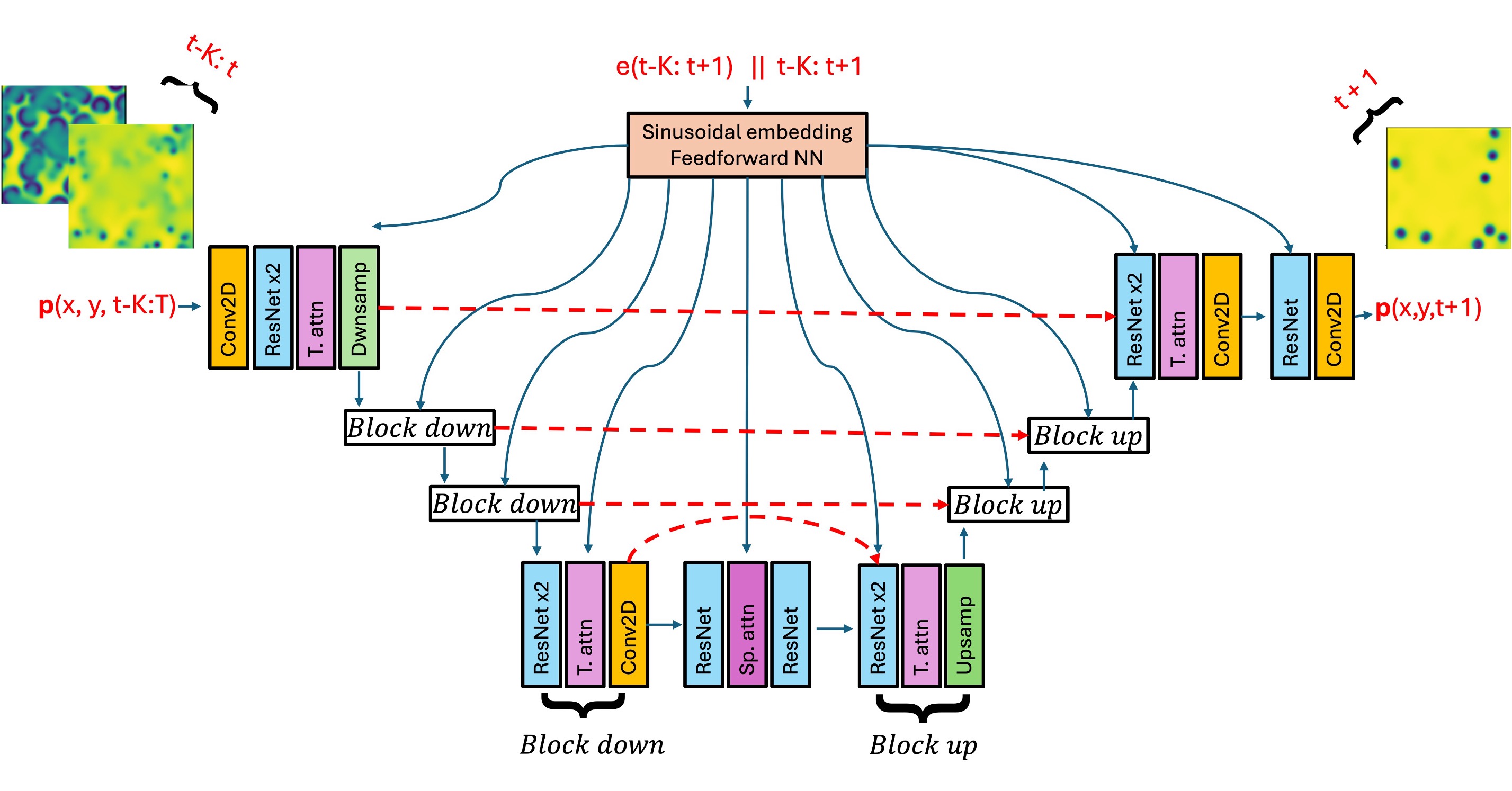}
    \caption{Schematic representation of the Diffusion-inspired Temporal Transformer Operator for spatiotemporal field prediction.}
    \label{fig:ditto}
\end{figure}

The training data, $\mathbf{D}_{train}$, is formed as tuples of the form: $\{\mathbf{p}(\mathbf{x},\mathbf{y},t_{k-(L-1):k}),$ $~e(t_{k-(L-1):k+1}), ~\mathbf{p}(\mathbf{x},\mathbf{y},t_{k+1})\}$
$\forall ~k \in (L,N_{T}-1)$. This creates a total of $N_{s}* (N_{T} - L)$ training samples, where $N_s$ is the total number of unique trajectories simulated (either with variation only in initial conditions, external time-dependent variable, or both). The model is trained by empirical risk minimization, 
\begin{equation}
    \mathcal{M}^{*} = \arg \min_{\mathcal{M} \in \mathbf{M}} \sum_{\mathbf{p}_i \in \mathbf{D}_{train}}\mathcal{L}(\mathcal{M}(\mathbf{p}_{i}) ,~\mathbf{p}_{i}), 
\end{equation}
where $\mathcal{M}^{*}$ is the choice of model parametrized by a set of weights within the entire model space $\mathbf{M}$ that minimizes an appropriate loss function $\mathcal{L}(\cdot)$ evaluated over all the training data $\mathbf{p}_{i} \in \mathbf{D}_{train}$. The mean squared error is chosen as the loss function in this work.

In this work, we utilize the Diffusion-inspired Temporal Transformer Operator (DiTTO) for temporal evolution of fields \cite{ovadia2023real}. In this approach, the time-dependent forcing function is encoded via a separate feedforward neural network (FNN) and this embedding is multiplied with each of the UNet intermediate layers, conditioning the UNet on the time-dependent variable. Each block in the UNet consists of a convolutional layer with a residual connection (ResNet block \cite{he2016deep}). This is followed by a temporal attention layer that captures dependencies across temporal features (physical fields at consecutive time instants represented as channels of an image). The attention mechanism \cite{vaswani2017attention} is denoted as,
\begin{equation}\label{eq:attn}
    \text{attn} = \text{softmax}(\frac{\mathbf{Q}\cdot\mathbf{K}^{T}}{\sqrt{d_{emb}}})\mathbf{V},
\end{equation}
where $\mathbf{Q}$, $\mathbf{K}$, $\mathbf{V}$, and $d_{emb}$ are the query, key, value, and the embedding dimensionality, respectively. The matrices $\mathbf{Q}$, $\mathbf{K}$, and $\mathbf{V}$ are essentially learnable representations of the input features obtained to extract self-dependencies across either the spatial or temporal dimensions of the physical fields. The embedding dimensionality normalizes the attention to avoid scaling issues and stabilize training. A detailed illustration of DiTTO architecture is shown in Fig.~\ref{fig:ditto}.

At inference time, the ground truth for the first $L$ time-steps is input to the model $\mathcal{M}$. The model predicts the field at the next time step. This is repeated recursively, with the model taking the $L$ recent model predictions to predict the successive time-step. At the end of the inference process, the entire trajectory prediction, $\widehat{\mathbf{p}}(\mathbf{x},\mathbf{y},t_{L+1:N_{T}})$,  is obtained. 

\begin{algorithm}
\caption{Ensemble Approach for Accurate Spatiotemporal Prediction.}
\label{alg:moe}
\begin{algorithmic}
\State \textbf{Training Phase:}
\State \textbf{Step 0:} Prepare the training dataset $\mathcal{D}_{\text{tr}}$ as a collection of tuples:
\[
\mathcal{D}_{\text{tr}} = \left\{ \left( \mathbf{p}(\mathbf{x}, \mathbf{y}, t_{k-L:k}),\ e(t_{k-L:k+1}),\ \mathbf{p}(\mathbf{x}, \mathbf{y}, t_{k+1}) \right) \right\} \quad \forall\ k \in [L, N_T - 1]
\]
\State \textbf{Step 1:} Train an ensemble of $N$ base models $\{\mathcal{M}_i\}_{i=1}^N$, each initialized with different random weights. Each model $\mathcal{M}_i$ is trained on the same dataset $\mathcal{D}_{\text{tr}}$ to learn the mapping from inputs to the next time step prediction.
\bigbreak
\State \textbf{Inference Phase:}
\State \textbf{Step 2:} Provide the ensemble with initial ground truth fields for the first $L$ time steps:
\[
\mathbf{p}(\mathbf{x}, \mathbf{y}, t_{1:L})
\]

\For{$k = L$ to $N_T - 1$}
    \For{$i = 1$ to $N$}
        \State Each model $\mathcal{M}_{i}$ predicts $\mathbf{p}(\mathbf{x}, \mathbf{y}, t_{k+1})$ from $\mathbf{p}(\mathbf{x}, \mathbf{y}, t_{k-L+1:k})$:
        \[
        \widehat{\mathbf{p}}_i(\mathbf{x}, \mathbf{y}, t_{k+1}) = \mathcal{M}_i\left( \mathbf{p}(\mathbf{x}, \mathbf{y}, t_{k-L+1:k}) \right)
        \]
    \EndFor
    \State Compute the ensemble-averaged prediction:
    \[
    \widehat{\mathbf{p}}_{\text{final}}(\mathbf{x}, \mathbf{y}, t_{k+1}) = \frac{1}{N} \sum_{i=1}^{N} \widehat{\mathbf{p}}_i(\mathbf{x}, \mathbf{y}, t_{k+1})
    \]
    \State Append $\widehat{\mathbf{p}}_{\text{final}}(\mathbf{x}, \mathbf{y}, t_{k+1})$ to the input sequence for the next prediction.
\EndFor

\State \textbf{Output:} The ensemble-generated spatiotemporal predictions:
\[
\left\{ \widehat{\mathbf{p}}_{\text{final}}(\mathbf{x}, \mathbf{y}, t_{L+1}), \widehat{\mathbf{p}}_{\text{final}}(\mathbf{x}, \mathbf{y}, t_{L+2}), \ldots, \widehat{\mathbf{p}}_{\text{final}}(\mathbf{x}, \mathbf{y}, t_{N_T}) \right\}
\]
\end{algorithmic}
\end{algorithm}

\subsection{Deep ensemble for autoregressive predictions}
\label{subsec:error_analysis}
In autoregressive models for time-dependent systems, the predicted output at one time step is fed back as input for the next. While effective, this approach introduces a compounding error mechanism - small inaccuracies at early steps can propagate and grow over time. One remedy is to use a deep ensemble approach: an ensemble of models, each trained independently on the same dataset but initialized with different random weights \cite{lakshminarayanan2017simple}. Rather than selecting a single “best” model, retaining all trained models in an ensemble offers several advantages: (i) different initializations explore distinct regions of the function space, resulting in diverse yet accurate approximations; (ii) error accumulation in autoregressive inference is mitigated through averaging; and (iii) no computational effort is wasted, as all trained models contribute to the final prediction.

Theoretical support for deep ensemble comprising of deep neural networks highlights their ability to capture multiple function modes due to random initialization and non-convex loss landscapes \cite{fort2019deep}. Previous work has explored ensembles in the context of neural operators for PDEs, using independently trained trunk networks to enhance generalization in static (single time-step) predictions \cite{sharma2024ensemble}. Building on this idea, we extend the deep ensemble paradigm to the dynamic setting, demonstrating its effectiveness for a range of spatiotemporal field evolution problems. Our approach is outlined in Algorithm~\ref{alg:moe}.

\indent To explain the error improvement using the deep ensemble approach, let each model prediction be written as,
\begin{equation}
    \mathcal{M}_{n}(\mathbf{z}) = f({\mathbf{z}}) + \epsilon_{n}({\mathbf{z}}),
\end{equation}
where $\mathcal{M}(\mathbf{z})$ is the output of the $n^{th}$ base model for a given input $\mathbf{z}$, $f(\mathbf{z})$ is the true output for $\mathbf{z}$, and $\epsilon_{n}(\mathbf{z})$ is the error by the $n^{th}$ base model for the input. The mean squared error for each model (MSE$_{n}$) is denoted as,
\begin{equation}\label{eq:mse}
    \mathbb{E}_{z}[\epsilon_{n}(\mathbf{z})^{2}] = \mathbb{E}_{z}[(\mathcal{M}_{n}(\mathbf{z}) - f(\mathbf{z}))^{2}],
\end{equation}
where $\mathbb{E}_{z}(\cdot)$ is the expectation operator. The MSE of the ensemble can be written as,
\begin{equation}\label{eq:mse_ens}
    \text{MSE}_{ens} = \mathbb{E}_{z}[(\frac{1}{N}\sum_{i=1}^{N}\mathcal{M}_{i}(\mathbf{z})- f(\mathbf{z}))^{2}],
\end{equation}
This expression can be decomposed into the correlated and uncorrelated components as
\begin{equation}\label{eq:mse_ens_dec}
    \text{MSE}_{ens} = \frac{1}{N^{2}}\sum_{i=1}^{N}\mathbb{E}_{z}[(\mathcal{M}_{i}(\mathbf{z}) - f(\mathbf{z}))^{2}] ~+~ \frac{1}{N^{2}}\sum_{i\neq j}\mathbb{E}_{z}[(\mathcal{M}_{i}(\mathbf{z})- f(\mathbf{z}))(\mathcal{M}_{j}(\mathbf{z})- f(\mathbf{z}))].
\end{equation}

\noindent Consider the hypothetical case that all model errors have zero mean (zero bias assumption) and are uncorrelated (independent). Then the ensemble error is only the first term in Eq.~\ref{eq:mse_ens_dec},
\begin{equation}\label{eq:mse_ens_lb}
    \text{MSE}_{ens}^{lower} = \frac{1}{N^{2}}\sum_{i=1}^{N}\mathbb{E}_{z}[(\mathcal{M}_{i}(\mathbf{z}) - f(\mathbf{z}))^{2}] = \frac{1}{N}(\frac{1}{N}\sum_{i=1}^{N}\text{MSE}_{i})
\end{equation}
which is exactly $1/N$ times the MSE averaged over all individual models. This is the theoretical lower bound on the ensemble error. Conversely, if all models are perfectly correlated (i.e., errors are identical), the ensemble MSE reduces to the mean of individual MSEs (see \cite{Brown2010} for proof):
\begin{equation}\label{eq:mse_ens_ub}
    \text{MSE}_{ens}^{upper} = \frac{1}{N}\sum_{i=1}^{N}\text{MSE}_{i}
\end{equation}
Eqns.~\ref{eq:mse_ens_lb} and \ref{eq:mse_ens_ub} together bound the ensemble MSE. Given that our approach has models with the same architecture and trained on the same data, it is unlikely that the models are uncorrelated. Hence, the total ensemble MSE shall also contain the term due to interaction between models as given in Eq.~\ref{eq:mse_ens_dec}. The diversity between different models (second term in Eq.~\ref{eq:mse_ens_dec}) determines the relative performance of the ensemble approach. This diversity depends on the training data, individual base models, and the overall optimization process, and analyzing that is beyond the scope of this work. Nevertheless, we demonstrate for a variety of spatiotemporal datasets, that the deep ensemble approach outperforms all the individual base models.

\section{Datasets}
In this work, we have demonstrated the deep ensemble approach on three high-dimensional PDEs. The details for all 3 datasets are summarized in Table.~\ref{table:dataset_details}. Each dataset is discussed in detail in the subsequent subsections. Each dataset has different spatiotemporal variations and hence presents a unique set of challenges.
\begin{table}[!h]
\centering
\caption{\normalsize{Details of the spatiotemporal datasets considered in this work}}
\vspace{1em}
\label{table:dataset_details}
\begin{tabular}{>{\arraybackslash} p{4.2cm} 
>{\centering\arraybackslash} p{2.0cm} 
>{\centering\arraybackslash} p{1.8cm} 
>{\centering\arraybackslash}  p{2.0cm} 
>{\centering\arraybackslash}  p{2.0cm} }
Dataset & Trajectories & Timesteps  & Input ~~~ field  & Output field  \leavevmode\\
 \hline
 \hline
 2-phase microstructure 
 \newline 
 (stress field: $\mathbf{\sigma_{xx}}$) & $120$ & $33$ & $\mathbf{\sigma_{xx}}$ history   + 
 \newline load path  & $\mathbf{\sigma_{xx}}$ future time-step
 \leavevmode\\
  \hline
 Gray-Scott diffusion \newline
 (species conc: $\mathbf{A}$) & $1200$ & $33$ & $\mathbf{A}$ history  & $\mathbf{A}$ future time-step
 \leavevmode\\
  \hline
  Shallow Water \newline Equation 
  \newline
  (velocity field: $\mathbf{u}_{x}$) & $120$ & $33$ &  $\mathbf{{u}}_{x}$ history & $\mathbf{u}_{x}$ future time-step
\leavevmode\\
 \hline
\end{tabular}
\end{table}

\subsection{Stress field prediction in a 2-Phase microstructure}
\begin{figure}[!h]
    \centering
    \includegraphics[width=0.99\linewidth]{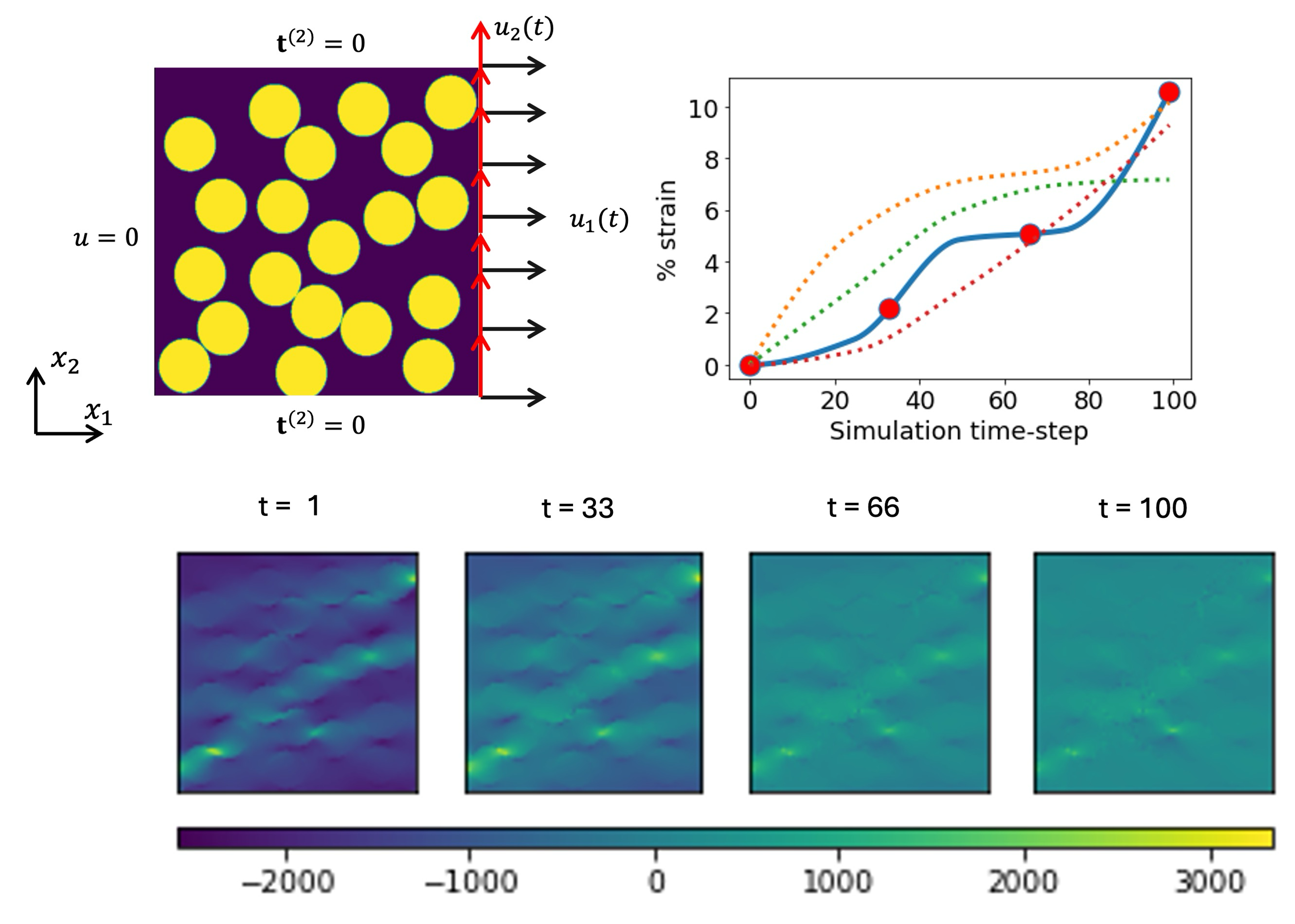}
    \caption{2-phase microstructure dataset: The top row illustrates a representative two-phase microstructure, the applied boundary conditions, and a sample load path generated using Algorithm~\ref{alg:loadpath_gen}. The bottom row shows the evolution of the stress tensor component ($\sigma_{xx}$) at selected time steps $t = {1, ~T/3, ~2T/3, ~T}$ with $T = 100$ for the load path shown in solid deep blue line.}
    \label{fig:microstructure_dataschematic}
\end{figure}

The first dataset comprises 2D two-phase microstructures subjected to time-dependent external loading (dynamic load paths), resulting in a spatiotemporally varying stress field. Spatial variation arises from the random placement of elastic fibers within the microstructure, while temporal variation is driven by the evolving external load. This setup, previously studied in \cite{saha2024prediction}, was used to validate a UNet-based model for quasi-static prediction of stress and strain under fixed load paths.

In this work, we consider multiple realizations of such microstructures, each containing 20 identical elastic fibers randomly embedded within an elastic perfectly-plastic matrix. The fibers, of constant radius, occupy 40\% of the microstructural volume. Each configuration is subjected to uniaxial ($u_1(t)$) and shear ($u_2(t)$) displacements on the right boundary, as illustrated in Fig.~\ref{fig:microstructure_dataschematic}. The left boundary is pinned, and the top and bottom edges are traction-free.

To simplify, both tensile and shear components follow randomly generated monotonically increasing load paths, generated as described in Algorithm~\ref{alg:loadpath_gen}. Each load path begins at zero strain, with the final strain sampled from a uniform distribution. A few intermediate time points are sampled to form a strictly increasing sequence, and a cubic interpolator \cite{fritsch1984method} is applied to obtain a smooth monotonic curve. This procedure is repeated to generate a diverse set of load paths, as illustrated in Fig.~\ref{fig:microstructure_dataschematic}.

\begin{algorithm}[!h]
\caption{Load Path Generation for 2-Phase Microstructure.}
\label{alg:loadpath_gen}
\begin{algorithmic}
\State \textbf{Initialize:}
\State \hspace{1em} Number of discrete time steps $T = 100$
\State \hspace{1em} Number of control points $N_{\text{ctrl}} = 5$
\State \hspace{1em} Initialize strain array $L \in \mathbb{R}^{T}$ with zeros
\State \hspace{1em} Initialize time array $X \in \mathbb{R}^{T}$ with zeros
\State \hspace{1em} Set $L[1] = 0$, $X[1] = 0$
\State \hspace{1em} Sample final strain: $L[T] \sim \mathcal{U}(0.07, 0.11)$
\State \hspace{1em} Set $X[T] = T$

\Statex

\State \textbf{Sample intermediate control points:}
\For{$i = 2$ to $N_{\text{ctrl}} - 1$}
    \State Sample $L[i] \sim \mathcal{U}(L[i-1], L[T])$ \Comment{Monotonic strain}
    \State Sample $X[i] \sim \mathcal{U}(X[i-1], X[T])$ \Comment{Monotonic time}
\EndFor

\Statex

\State \textbf{Fit interpolator:}
\State Fit a monotonic cubic interpolator (e.g., Fritsch–Carlson) to \\ $\{X[1:N_{\text{ctrl}}], L[1:N_{\text{ctrl}}]\}$
\State Evaluate interpolator to obtain full strain trajectory $L[1:T]$
\end{algorithmic}
\end{algorithm}

Stress distributions are calculated using the ABAQUS finite element software \cite{barbero2023finite}, assuming a $J_2$ plasticity model without hardening, where the yield function is defined as,
\begin{equation}
    f(\mathbf{S}) = \sqrt{3J_2}-\sigma_{y},
\end{equation}
and $J_2$ is defined as,
\begin{equation}
    J_2 = \frac{1}{2}\mathbf{S : S}
\end{equation}
where $\mathbf{S}$ is the deviatoric part of the Cauchy stress tensor and $\sigma_{y}$ = $210$ MPa is the yield strength of the material. The Young's modulus for the matrix and fiber are set to 3.2 GPa and 87 GPa, respectively. The Poisson's ratio for the matrix and fiber are set to $0.35$ and $0.20$, respectively. Additionally, the model assumes small-strain conditions, which allows the additive decomposition of the strain tensor into elastic and plastic components as shown in Eq.~\ref{eq: additive decomposition}.
\begin{equation}
    \varepsilon =  \varepsilon_e + \varepsilon_p
    \label{eq: additive decomposition}
\end{equation}

The simulations use an irregular mesh with 86000 elements on average to resolve the smaller regions between two fibers. The full dataset can be found at \cite{data_microstructure}. A mesh convergence study of this finite element model can be found in ~\cite{saha2024prediction}. After the simulation, the stress fields obtained are interpolated on a regular square grid ($N_{x} \times N_{y}$ = $128 \times 128$) for implementation in the ML model. For model training, the first time instant ($t=0$) corresponding to no external strain is discarded and then we subsample the load path by 3 (i.e., $N_{T} = 33$).  

One example of a generated load path and its corresponding stress field simulation (with dimensions $N_x \times N_y \times N_T$) is illustrated in Fig.~\ref{fig:microstructure_dataschematic}, where intermediate time steps are uniformly spaced. The time axis is normalized, i.e., each load path is represented by $T$ discrete values between 0 and 1, which are then used as input to the simulation model. The solid blue curve indicates the specific load path for which the spatial stress field evolution is visualized. Stress maps at selected time instances $t = {1, ~T/3, ~2T/3, ~T}$, marked by red circles, are shown alongside the load path. For training and testing the model, the stress field is subsampled in time by a factor of 3, resulting in $N_t = 33$ total time steps. 
Obtaining full-field stress data from ABAQUS for a single time-step requires $1.6$s, while the inference time for the deep ensemble approach is $9$x faster. In this work, we focus on predicting the evolution of a single stress component ($\mathbf{\sigma_{xx}}$); however, the framework can be easily extended to include other stress tensor components. 

\subsection{Gray-Scott reaction diffusion}

\begin{figure}[!h]
    \centering
    \includegraphics[width=0.99\linewidth]{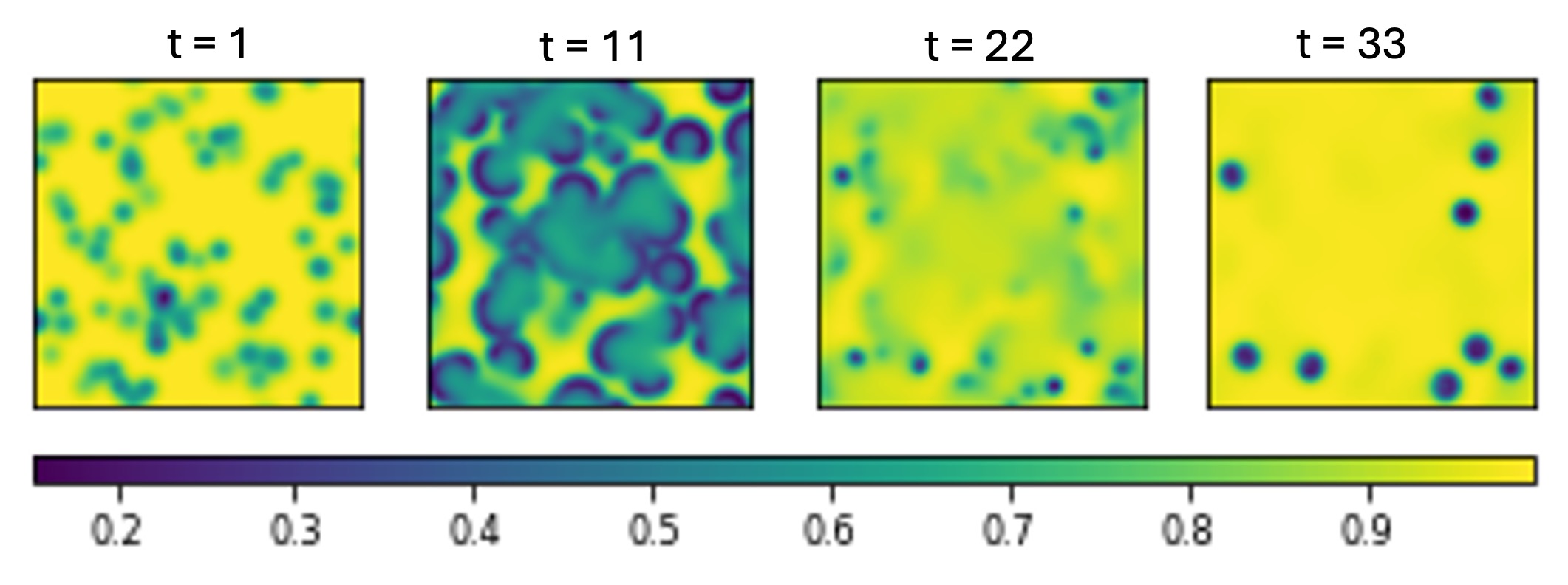}
    \caption{Gray-Scott reaction diffusion dataset: Species A concentration shown for a random initial condition and Gray-Scott parameters $F = 0.098$, $K = 0.057$ at time steps $t = {1, ~T/3, ~2T/3, ~T}$, with $T = 33$ on the simulation trajectory.}
    \label{fig:grayscott_sim}
\end{figure}

The Gray-Scott reaction diffusion model \cite{gray1984autocatalytic} describes systems thatare commonly found in many biological and chemical species where chemical reactions interact with a spatial diffusion process. Let us consider two chemical species $A$ and $B$ interacting with each other in a closed domain. The Gray-Scott reaction diffusion model describes the variation of chemical concentration of both species over space and time as:
\begin{align}
    \frac{\partial A}{\partial t} &= \delta_{A}\Delta A - AB^{2} + f(1 - A), \label{eq:grayscott_pde1} \\
    \frac{\partial B}{\partial t} &= \delta_{B}\Delta B + AB^{2} - (f + k)B. \label{eq:grayscott_pde2}
\end{align}
In the above system, $f$ and $k$ are the feed and kill parameters, respectively. Specifically, $f$ controls the rate at which species $A$ is added to the system, and $k$ controls the rate at which species $B$ is removed from the system.   $\delta_{A}$ and $\delta_{B}$ control the diffusion rate of each species and are set to $2 \times 10^{-5}$ and $1 \times 10 ^{-5}$, respectively, for the simulations.

The simulation data is obtained from \cite{ohana2025well}. Eqs.~\ref{eq:grayscott_pde1} - ~\ref{eq:grayscott_pde2} are simulated with a higher order spectral difference method on the spatial domain $[-1, -1] \times [1, 1]$ with periodic boundary conditions and $\Delta t = 10s$. At each time step, the simulation is represented as a regular grid of size $128 \times 128$, with 200 initial conditions (random Gaussian clusters), and 6 different combinations of $f$ and $k$ parameters are considered for the simulations. This leads to a diverse dataset ($200 \times 6 = 1200$ unique patterns). The details of the 6 combinations of $f$ and $k$ parameters are detailed in Table.~\ref{table:grayscott-simdetails}.
We demonstrate our approach for predicting the evolution of the species A chemical concentration for the first $N_{t} = 33$ simulation timesteps.

\begin{table}[!h]
\centering
\caption{Parameter combinations considered for the dataset generation of the Gray-Scott reaction diffusion system.}
\label{table:grayscott-simdetails}
\begin{tabular}{>{\centering\arraybackslash} p{1.5cm} 
>{\centering\arraybackslash} p{1.5cm} 
>{\centering\arraybackslash} p{1.5cm}}
F & K & Pattern  \leavevmode\\
 \hline
 \hline
$0.014$ & $0.054$ & Gliders 
 \leavevmode\\
  \hline
$0.098$ & $0.057$ & Bubbles
 \leavevmode\\
  \hline
$0.029$ & $0.057$ & Maze 
 \leavevmode\\
  \hline
$0.03$ & $0.062$ & Spots 
 \leavevmode\\
  \hline
  $0.058$ & $0.065$ & Worms 
 \leavevmode\\
  \hline
$0.018$ & $0.051$ & Spirals 
 \leavevmode\\
  \hline
 \hline
\end{tabular}
\end{table}

\subsection{Shallow Water Equation}

The shallow water equations are a 2D approximation of a 3D incompressible fluid flow when horizontal length scales are much larger than vertical length scales. They are obtained by numerically depth integrating the incompressible Navier-Stokes equation. These equations are crucial to validate atmospheric dynamics \cite{williamson1992standard}. Specifically, we consider the rotating forced hyperviscous spherical shallow water equation,
\begin{equation}
    \frac{\partial \mathbf{u}}{\partial t} = -\mathbf{u}\cdot\nabla\mathbf{u} - g\nabla h - \nu\nabla^{4}\mathbf{u} - 2\Omega \times\mathbf{u}, 
\end{equation}
\begin{equation}
 \frac{\partial h}{\partial t} = -H\nabla\cdot\mathbf{u} -\nabla \cdot(h\mathbf{u}) -\nu\nabla^{4}h + F,
\end{equation}
where \textbf{u} is the velocity vector field, $h$ is the deviation of pressure surface height from the mean, $H$ is the mean height, $\Omega$ is the Coriolis factor, and F is the forcing term to induce double periodicity (daily and annual). The $\nabla^{4}$ term corresponds to the hyperviscosity term. Spherical boundary conditions and periodic forcing term are adopted in this simulation to create a simpler proxy for real-world atmospheric dynamics. The initial conditions are sampled from hPa $500$ level in ERA5 \cite{hersbach2020era5}.

\begin{figure}[!h]
    \centering
    \includegraphics[width=0.99\linewidth]{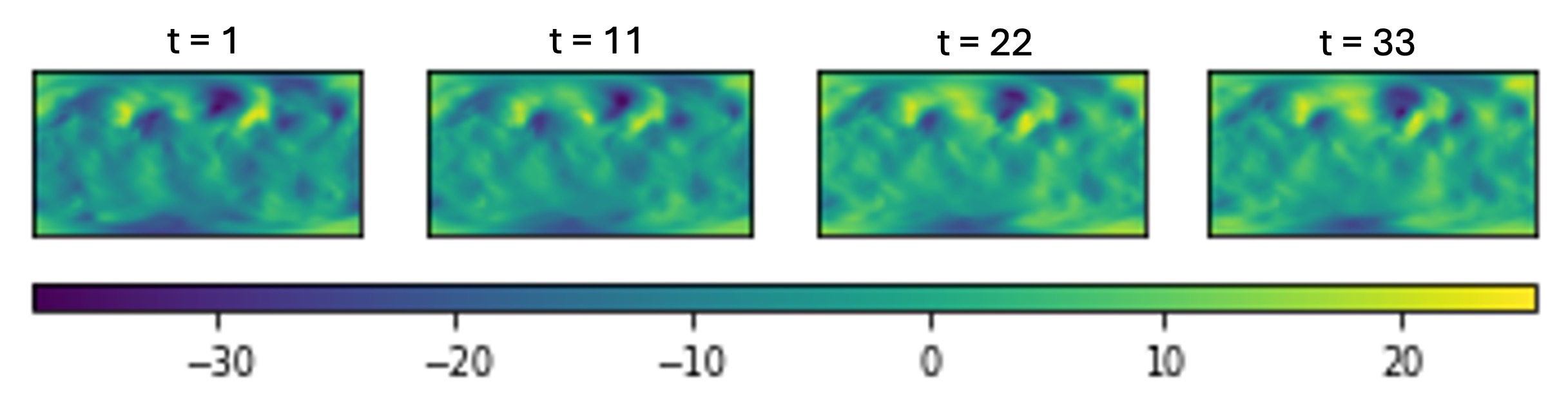}
    \caption{Shallow-water equation results (Shallow Water equation dataset): The velocity field component \( u_x \) is visualized at four key time instants along the simulation trajectory - $t = \{1, ~\frac{T}{3}, ~\frac{2T}{3}, ~T\}$, with $T = 33$. These snapshots capture the spatiotemporal evolution of the flow dynamics throughout the simulation.}
    \label{fig:planetswe_sim}
\end{figure}

The dataset used in this work has been previously introduced in an atmospherical dynamics prediction study using Fourier Neural Operator (FNO)\cite{mccabe2023towards} and a comprehensive dataset study \cite{ohana2025well}. The dataset consists of $120$ trajectories with $\Delta t = 1$ hour in simulation time. The data is simulated on an equiangular grid with $\phi \in [0, 2\pi]$ and $\theta \in [0, \pi]$. At each time-step, the high-dimensional simulation is represented as a $256 \times 512$ 2D image corresponding to the equiangular spatial domain. We demonstrate our approach for the evolution of the velocity vector field component ${u}_{x}$ for the first $N_{t} = 33$ simulation timesteps.

\subsection{Temporal information processing}
As explained in Sec.~\ref{sec:methods}, the base models in the deep ensemble are conditioned on time via an embedding of the temporal information. For each dataset, the temporal input has two parts: time-IDs and the external forcing input. For the 2-phase microstructure dataset, the temporal input is a 1D vector of size $2(L+1)$, consisting of the external forcing input values $e(t_{k-(L-1):k+1})$ - that is, the load values over the previous $L$ time steps and the next time step, along with the corresponding discrete time IDs $(t_{k-(L-1):k+1})$, as illustrated in Fig.~\ref{fig:ditto}.

For the Gray-Scott reaction-diffusion dataset, the temporal input is a 1D vector of size $(L+1)+2$, comprising the previous $L$ time IDs plus the next one and the external forcing input (comprising of parameters $f$ and $k$ values from Eqs.~\ref{eq:grayscott_pde1}--\ref{eq:grayscott_pde2}). Note that while $f$ and $k$ are constant over time, their combination with different initial conditions leads to a distinct set of patterns.

For the Shallow Water equation dataset, the temporal input is a 1D vector of size $(L+1)$, representing the previous $L$ and the next time IDs. There is no external forcing input in this case, as the only source of variability is due to differences in initial conditions.

The final temporal embedding $u(t)$ is defined as:
\begin{equation}\label{eq:time_embedding_full}
u(t) = \text{FNN}_{\text{emb}}(t) \,||\, \text{FNN}_{\text{var}}(e(t))),
\end{equation}
where $\text{FNN}_{\text{emb}}(t)$ is the time-ID embedding, $\text{FNN}_{\text{var}}(e(t))$ is the forcing input embedding, and $||$ denotes vector concatenation.

The component $\text{FNN}_{\text{emb}}(t)$ encodes time IDs using a sinusoidal positional embedding - used in NLP to represent token positions-followed by two fully connected layers with Gaussian Error Linear Unit (GeLU) activation. The resulting output is a 1D vector of size $1 \times 32$ (output dimension obtained after hyperparameter tuning). The sinusoidal embedding is computed as:
\begin{equation}
s_i(t) = \sin\left(\frac{t}{10000^{i/d_t}}\right) \,||\, \cos\left(\frac{t}{10000^{i/d_t}}\right), \quad \text{for } i = 1, \dots, d_t,
\end{equation}
where $d_t$ is the embedding dimension.

The external forcing input $e(t)$ is embedded using $\text{FNN}_{\text{var}}(e(t))$, which consists of two linear layers with a GeLU activation in between. The output is also a 1D vector of size $1 \times 32$ (output dimension obtained after hyperparameter tuning). As described in Sec.~\ref{sec:methods}, the complete temporal embedding $u(t)$ is injected into each intermediate layer of the UNet. This is achieved by reshaping $u(t)$ and passing it through a 2D convolutional layer, ensuring the number of filters matches the channel size of the intermediate UNet output. For the Shallow Water equation dataset, which lacks time-varying external inputs, the temporal embedding simplifies to only the time-ID embedding (FNN$_\text{emb}(t)$).

\subsection{Deep Ensemble training}

For the 2-phase microstructure dataset, the time-varying load paths differ between the training and testing datasets. In contrast, for the Gray-Scott and Shallow Water equation datasets, the variability arises from different initial conditions in training and testing data. The training datasets are standardized for faster convergence during training. The test data are standardized with the training data statistics.

During training, we use a batch size of 8. At test time, predictions are generated in an autoregressive manner to reconstruct the full trajectory. Accordingly, the test-time batch size is set to $N_t - L$, where $N_t$ is the trajectory length for the given dataset and $L$ is the temporal history length. We set $L = 3$ (number of previous time-steps to be used as model input) for all datasets. In general, we recommend choosing $3 \leq L \leq 10$ to balance the trade-off between predictive accuracy and the amount of ground truth required at test time.

To mitigate overfitting, we apply $L_2$-norm regularization to the model weights during training. All base models are trained for 100 epochs using the Adam optimizer with a learning rate of $0.0001$ and default hyperparameters. Our primary goal is not to optimize performance for each individual dataset, but rather to demonstrate the effectiveness of the ensemble approach compared to standalone models. Training for 100 epochs is sufficient to ensure convergence to a stable optimum.

It is noted that for a fair comparison of numerical solvers with data-driven surrogate models, the model training time should also be accounted for. Since the deep ensemble models are randomly initialized and do not share any parameters, they can be trained/used in parallel on multiple GPUs. In this work, the models are trained on a GPU cluster node (each node has 4 Nvidia A100 GPUs with 40GB of memory each). For the 3 datasets, the average training time on GPU is $\approx 4.5$ hours for one model. After training, the models can be used at inference time for every new test sample without any retraining.

\section{Results}
The performance of the deep ensemble approach for all 3 datasets is compared qualitatively by visualizing the physical field predictions at select time-steps on the simulation trajectory. A quantitative comparison of the deep ensemble approach and individual models is presented next. The relative $\mathcal{L}^{2}$ norm error (RLE) is defined as, 
\begin{equation}
    \text{RLE}(\mathbf{p}(\mathbf{x},\mathbf{y},t), ~\widehat{\mathbf{p}}(\mathbf{x},\mathbf{y},t)) = \frac{1}{N_{test}}\sum_{i=1}^{N_{test}}\frac{\left \lVert  \mathbf{p}_{i}(\mathbf{x},\mathbf{y},t) - \widehat{\mathbf{p}}_{i}(\mathbf{x},\mathbf{y},t)\right \rVert_2}{\left \lVert  \mathbf{p}_{i}(\mathbf{x},\mathbf{y},t)\right \rVert_2}
\end{equation}

\noindent The mean absolute error (MAE) is defined as,
\begin{equation}
    \text{MAE}(\mathbf{p}(\mathbf{x},\mathbf{y},t), ~\widehat{\mathbf{p}}(\mathbf{x},\mathbf{y},t)) = \frac{1}{N_{test}}\sum_{i=1}^{N_{test}}\frac{1}{N_{x}*N_{y}}\sum_{x,y}  | p_{i}(\mathbf{x},\mathbf{y},t) - \widehat{p}_{i}(\mathbf{x},\mathbf{y},t) |,
\end{equation}
where $N_{test}$ is the size of the test dataset. Both metrics are computed separately at each time step $t$ since we are interested in evaluating the performance of the approaches over time.

\begin{figure}[!h]
    \centering
    \includegraphics[width=0.97\linewidth]{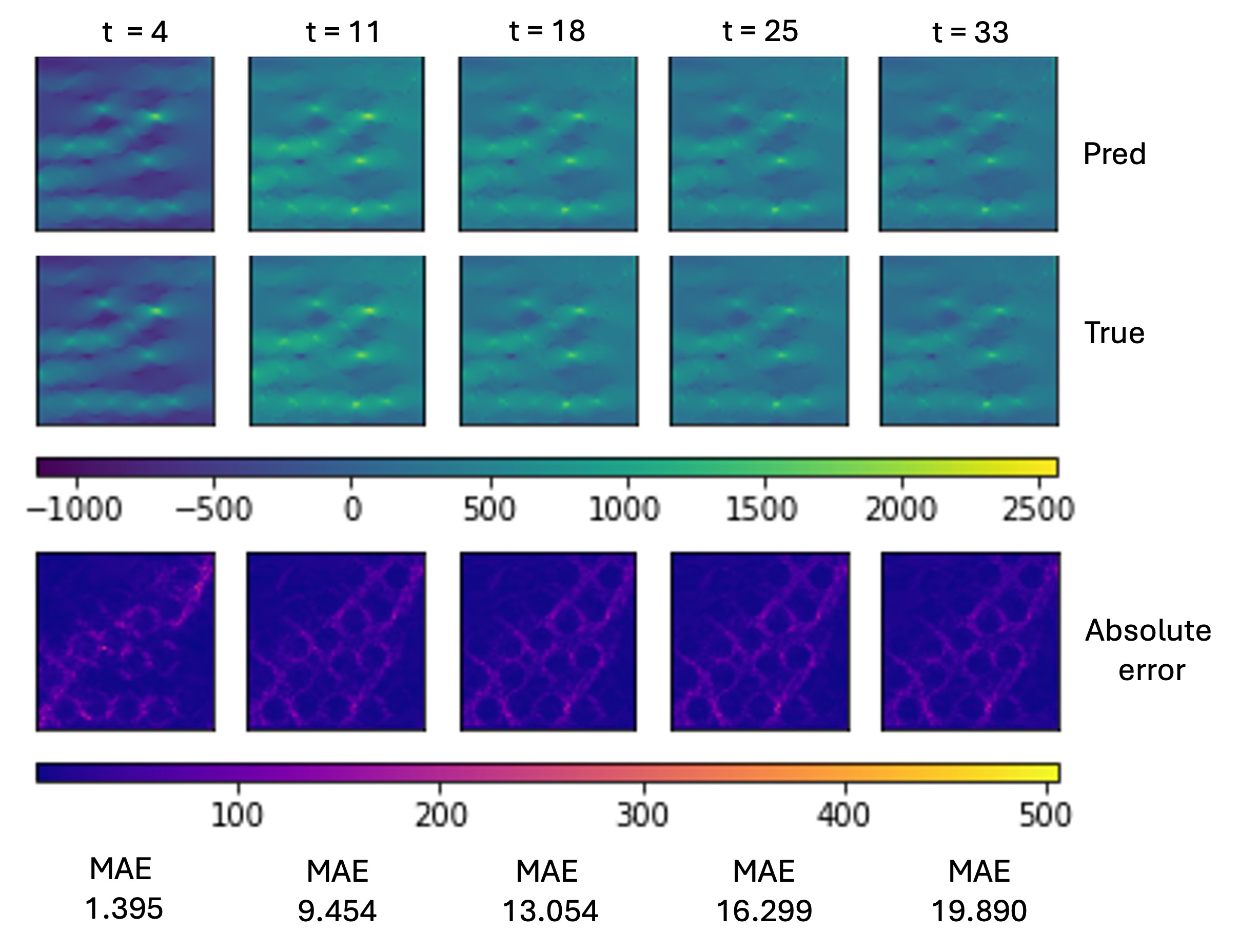}
    \caption{2-phase microstructure stress field prediction with the deep ensemble (top), reference FEM solution (middle row), and absolute error map (bottom) at t = $\{L+1, \frac{T}{4}, \frac{T}{2}, T\}$ time steps of the simulation trajectory.}
    \label{fig:micro_preds}
\end{figure}

\begin{figure}[!h]
    \centering
    \includegraphics[width=0.97\linewidth]{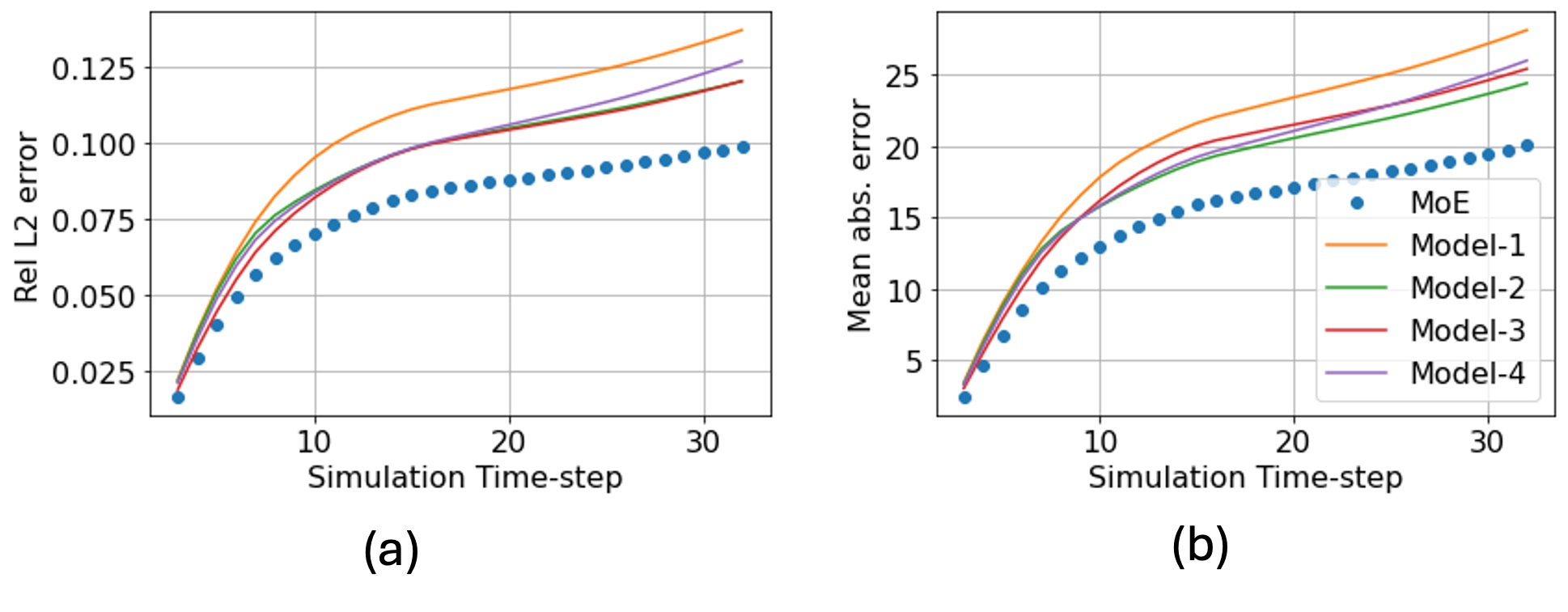}
    \caption{Performance metrics for the 2-phase microstructure dataset (a). RLE and (b). MAE.}
    \label{fig:metrics_micro}
\end{figure}

\subsection{2-phase microstructure}
Fig.~\ref{fig:micro_preds} shows the predictions with the deep ensemble approach at selected time-steps along the simulation trajectory. There is a remarkable similarity between the ground truth simulations (reference finite element solution) and the ensemble predictions. The absolute error is low except at a few points on the fiber-matrix interfaces.

The performance metrics with the deep ensemble approach are illustrated in Fig.~\ref{fig:metrics_micro}. As illustrated in Fig.~\ref{fig:metrics_micro}(a), at the end of the simulation trajectory, RLE metric with the deep ensemble approach is $0.0987$, which is $18$ to $28\%$ less than the RLE range with individual models ($0.120$ to $0.137$). Similarly, Fig.~\ref{fig:metrics_micro}(b) demonstrates that at the end of the simulation trajectory, the MAE metric is at least $20\%$ less than the best-performing individual model (Model 2 in this case). Both comparisons show that the deep ensemble approach helps to mitigate error accumulation in an autoregressive setting.

\begin{figure}[!h]
    \centering
    \includegraphics[width=0.97\linewidth]{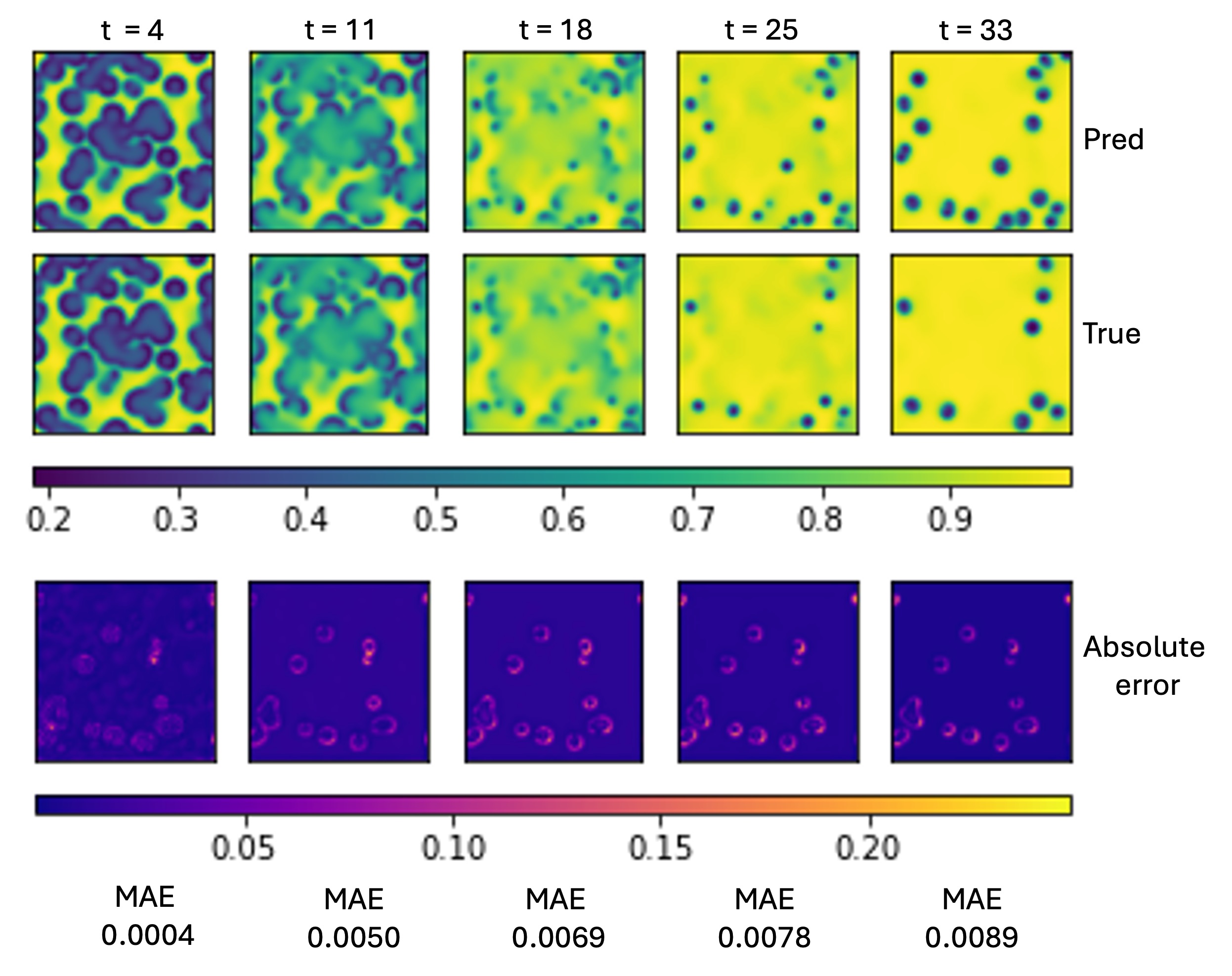}
    \caption{Gray-Scott reaction species concentration with the deep ensemble (top row), reference analytical solution (middle row), and absolute error map (bottom row) at t = $\{L+1, \frac{T}{4}, \frac{T}{2}, T\}$ time steps of the simulation trajectory.}
    \label{fig:grayscott_preds}
\end{figure}

\begin{figure}[!h]
    \centering
    \includegraphics[width=0.97\linewidth]{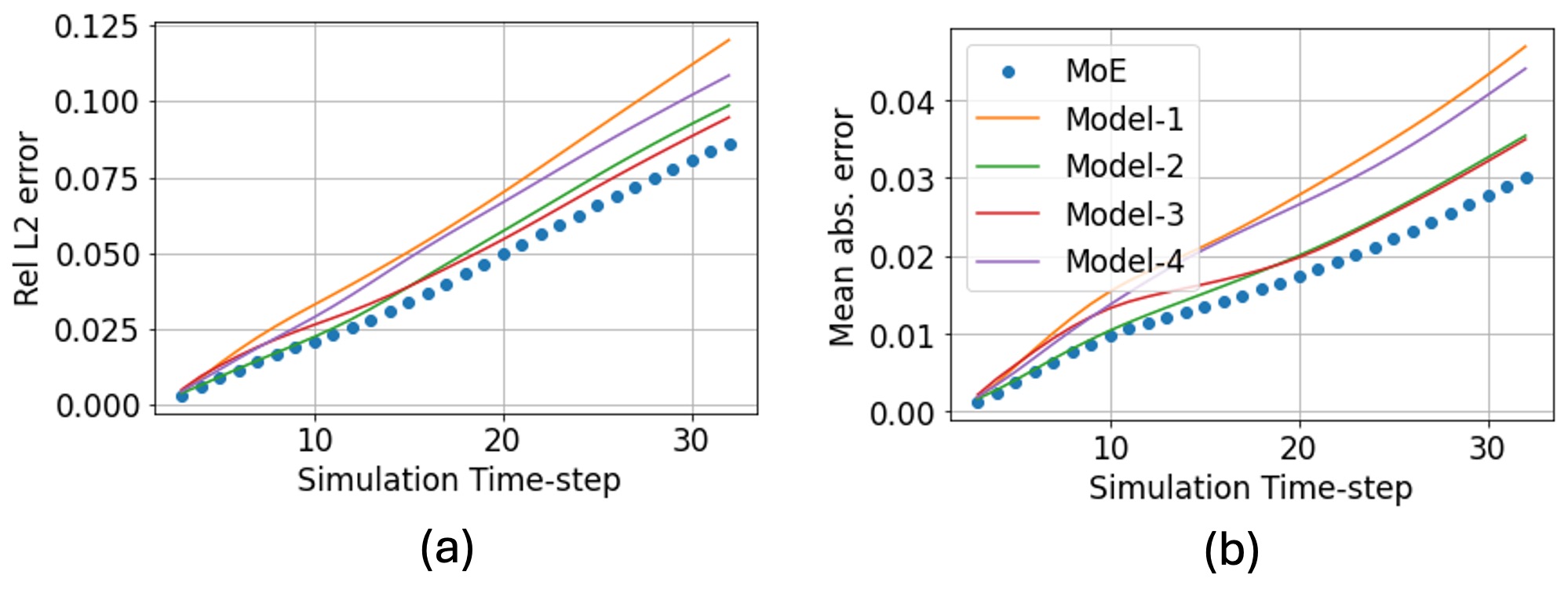}
    \caption{Performance metrics for the Gray-Scott diffusion reaction dataset (a). RLE and (b). MAE.}
    \label{fig:metrics_grayscott}
\end{figure}

\subsection{Gray-Scott reaction diffusion}
The results for the Gray-Scott diffusion reaction dataset, illustrated in  (Fig.~\ref{fig:grayscott_preds}), show more significant errors in the prediction. Specifically, a few of the patterns are mispredicted by the data-driven model at later timesteps (t=$25$ onwards), with the data-driven model predicting regions of very low species concentration that are not present in the ground-truth data. The deep ensemble approach improves the model predictions, with the RLE at the end of the trajectory being $0.085$, which is $5$ to $ 30\%$ less than the RLE metric range with individual models ($0.095$ to $0.12$), see Fig.~\ref{fig:metrics_grayscott}(a). Similarly, the MAE metric with the deep ensemble approach is at least $14\%$ less compared to an individual model (best performing Model-3), as illustrated in Fig.~\ref{fig:metrics_grayscott}(b).

\begin{figure}[!h]
    \centering
    \includegraphics[width=0.97\linewidth]{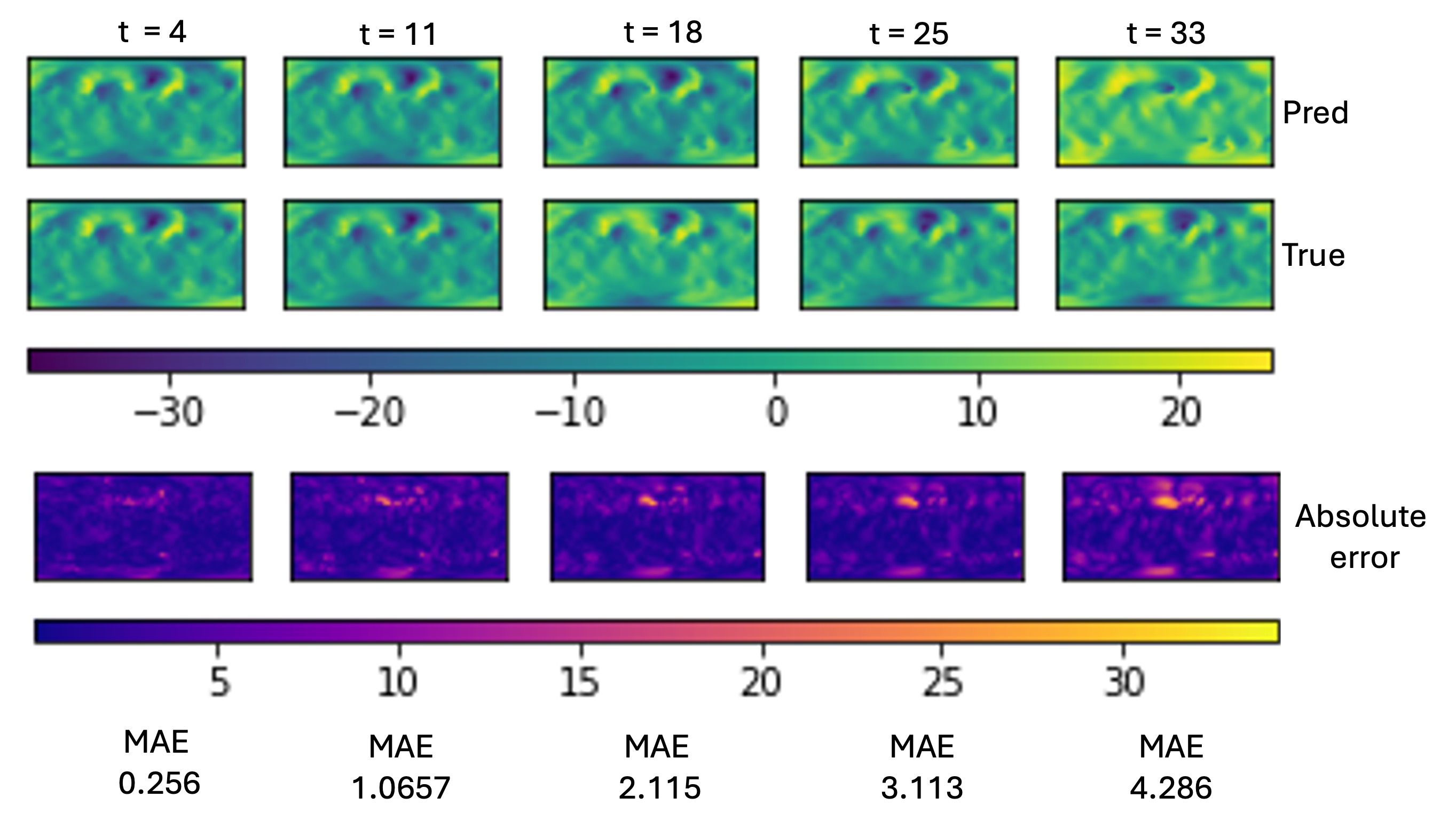}
    \vspace{5mm}
    \caption{Shallow Water equation velocity field prediction with the deep ensemble (top row), reference FEM solution (middle row), and absolute error map (bottom row) at t = $\{L+1, \frac{T}{4}, \frac{T}{2}, T\}$ time steps of the simulation trajectory with $L = 3, T = 33$ being the time history input to the model and total simulation steps respectively.}
    \label{fig:planetswe_preds}
\end{figure}

\subsection{Shallow water equation}
Fig.~\ref{fig:planetswe_preds} shows the predictions with the deep ensemble approach for the Shallow Water equation dataset. This illustration shows higher error in regions with complicated patterns (top center of the spatial dimensions). Towards the end of the simulation trajectory, the data-driven model is unable to capture all the patterns in the data. Indeed, researchers have demonstrated the inability of data-driven surrogate models to capture high-frequency patterns in physical systems \cite{wang2022and} (spectral bias). Note that it is inherently impossible for a data-driven surrogate model to capture all the subtleties of the physical system. The objective of this work is to showcase the advantage of using the deep ensemble instead of individual models.

\begin{figure}[!t]
    \centering
    \includegraphics[width=0.97\linewidth]{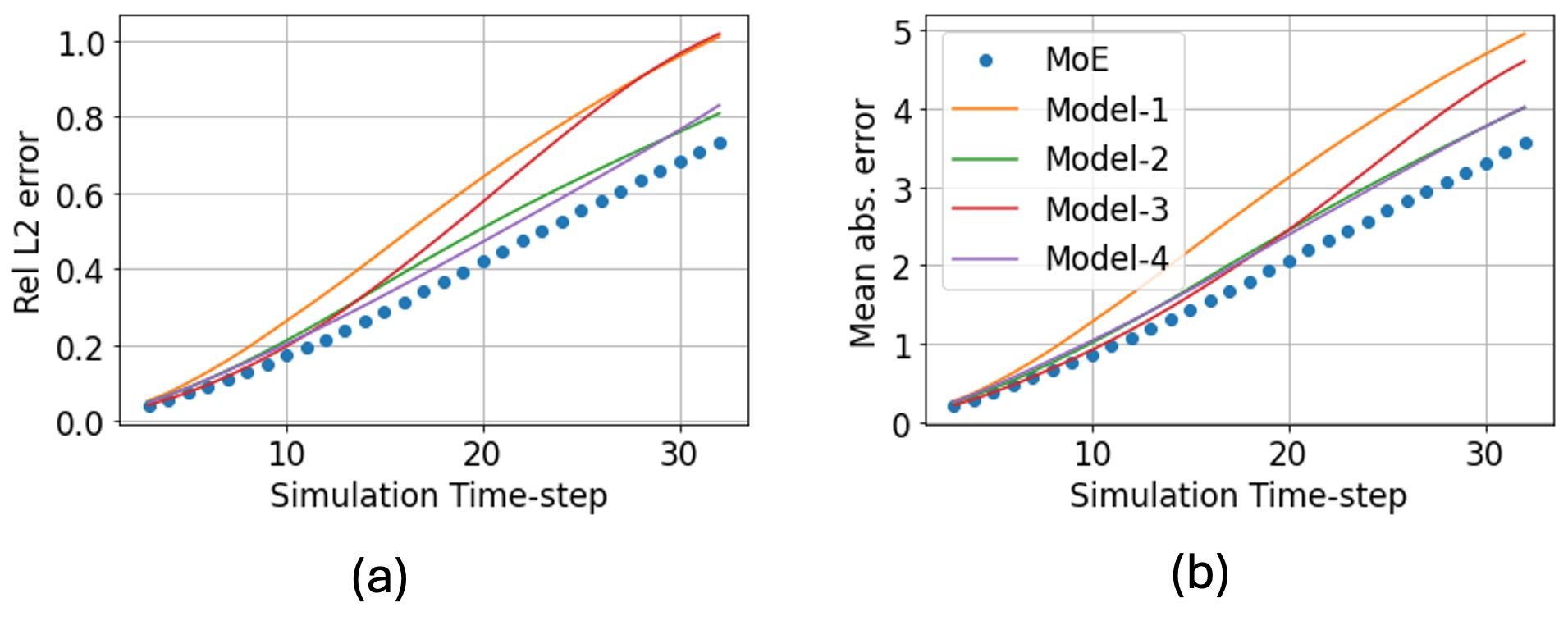}
    \caption{Performance metrics for the Shallow Water equation dataset (a). RLE and (b). MAE.}
    \label{fig:metrics_planetswe}
\end{figure}

Due to the complicated nature of the dataset, we expect the error metrics to be higher. As illustrated in Fig.~\ref{fig:metrics_planetswe}(a), with the deep ensemble approach, the RLE at the end of the simulation trajectory is $0.73$, which is $10$ to $ 30\%$ less than the RLE range with individual models ($0.81$ to $1.018$). The MAE metric with the deep ensemble approach is at least $15\%$ less compared to the individual models (best performing model 4), as illustrated in Fig.~\ref{fig:metrics_planetswe}(b).

\subsection{Analysis of the deep ensemble framework}

\begin{figure}[!h]
    \centering
    \includegraphics[width=0.97\linewidth]{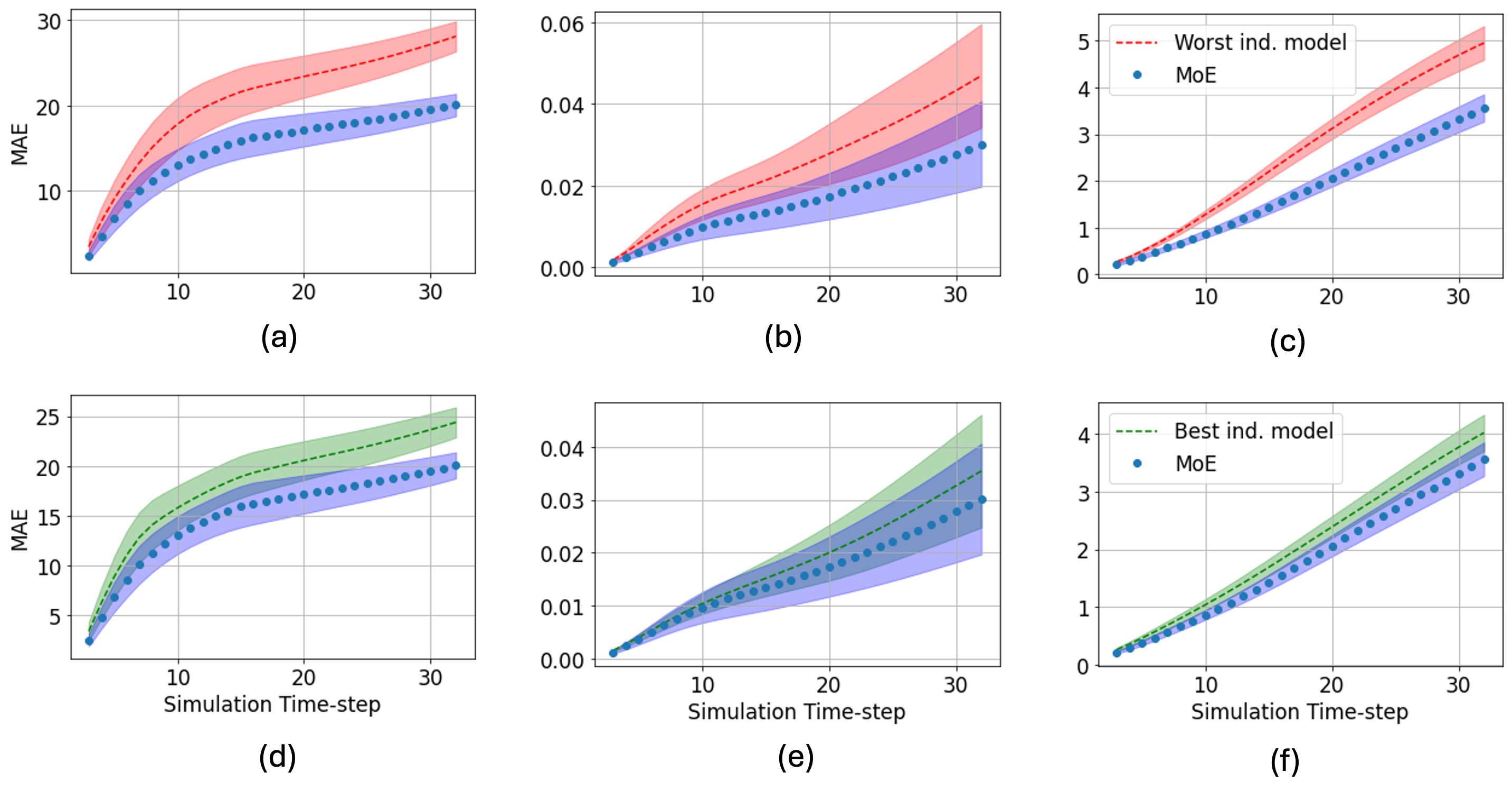}
    
    \caption{Performance comparison of the deep ensemble approach and model with highest MAE and lowest MAE respectively for: 2-phase microstructure dataset (a, d); Gray Scott diffusion reaction dataset (b, e); Shallow Water equation dataset (c, f). }
    \label{fig:bestworst_mae}
\end{figure}

As noted in Sec .~\ref {subsec:error_analysis}, the ensemble error decreases as the correlation between individual model errors decreases. 
Given that all the base models share the same architecture and parameters, the difference in performance can be attributed to the complexity of the individual datasets. This likely gives rise to different degrees of independence between individual models. 

Fig.~\ref{fig:bestworst_mae} compares the best and worst performing models v/s the deep ensemble approach for all three datasets (in terms of MAE metric). On average over all 3 datasets, the deep ensemble approach has a $33\%$ lesser MAE compared to the worst-performing individual model and $15\%$ lesser MAE compared to the best-performing model. This comparison underscores the observation that the deep ensemble approach always outperforms the best individual model and, on average, provides substantial improvement in performance.

\begin{figure}[h!]
    \centering
    \includegraphics[width=\textwidth]{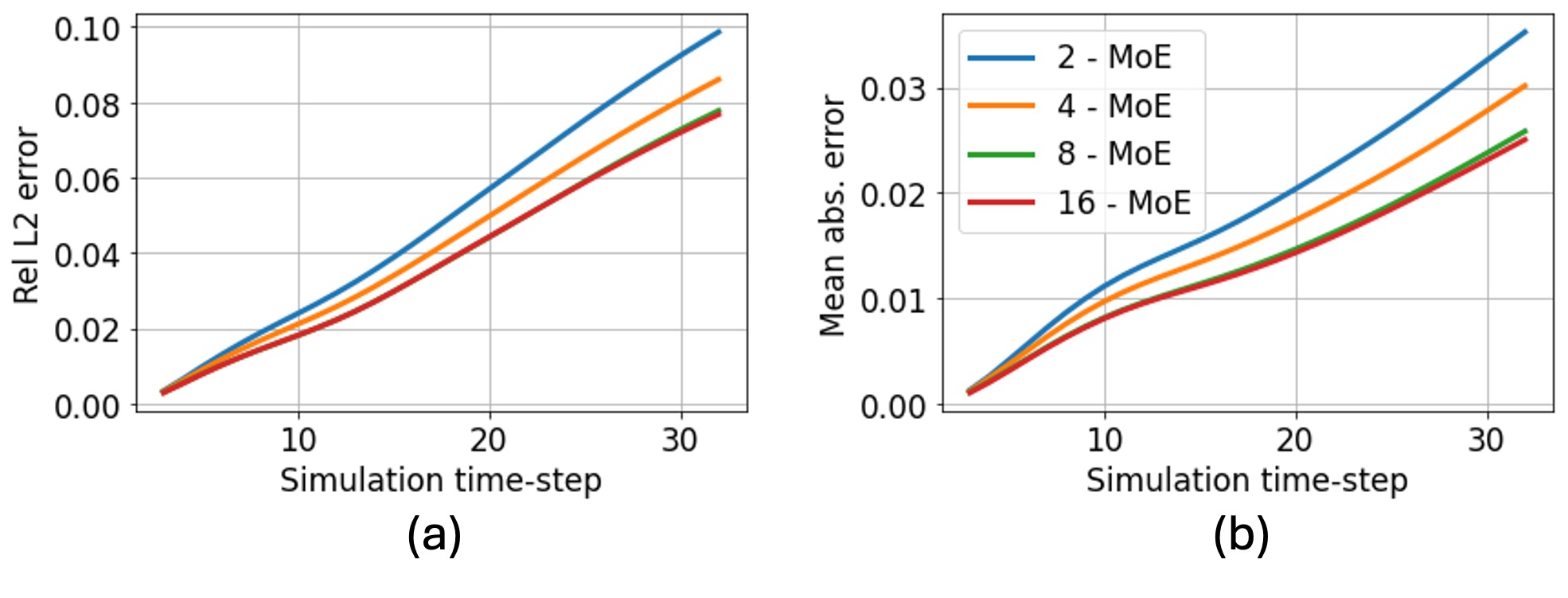}
    \caption{Performance comparison for different numbers of base models in the ensemble approach for Gray-Scott diffusion reaction dataset: (a). RLE and (b). MAE.}
    \label{fig:ensemble_nvar}
\end{figure}

Clearly, the results using the deep ensemble approach will improve on average, as the number of base models in the ensemble is increased, but the requisite computational effort will increase. To investigate this effect, Fig.~\ref{fig:ensemble_nvar} illustrates the performance metrics for a varying number of base models in the deep ensemble for the Gray-Scott reaction diffusion dataset. We observe that an increasing number of base models leads to improved performance (increasing the number of models from 2 to 8 improves the MAE by $28\%$). However, after a saturation point, further increases in the number of models ($8$ to $16$) do not lead to substantial improvement in performance. The number of base models in the deep ensemble has to be chosen based on a tradeoff between improvement in performance and time and memory requirement to train and store all base models respectively.

\section{Conclusions}
\label{sec:conclusion}

Traditional numerical solvers for time-dependent PDEs provide high-fidelity results but are often too computationally intensive for real-time or many-query applications. While ML surrogates offer a scalable alternative, autoregressive implementations are prone to error accumulation over long horizons-compromising their reliability in practical deployments.

In this work, we demonstrate that a simple ensemble strategy-based solely on random weight initialization-can significantly reduce long-term prediction error in autoregressive ML surrogates. By training multiple models on the same data with different initializations and averaging their predictions, we effectively mitigate the compounding errors typical of single-model rollouts. Across three diverse PDE systems-elastic stress evolution, shallow water dynamics, and reaction-diffusion chemistry-our ensemble approach improves predictive stability by 15–33\%, without requiring any architectural modifications, memory modules, or retraining overhead. The effectiveness of our method stems from its ability to harness the stochasticity of neural network training. Each model in the ensemble converges to a different local minimum, and their aggregated predictions offer improved generalization and robustness to initialization-induced bias. The result is a lightweight, architecture-agnostic enhancement that retains the computational speed of ML surrogates, making it particularly attractive for real-time simulation in scientific and engineering workflows.

Looking forward, our ensembling framework opens up several directions for future work. One avenue is to integrate physics-informed constraints or inductive biases to further regularize ensemble outputs. Another promising direction is to explore ensemble diversity not just through random initialization but also via controlled variations in hyperparameters, architectures, or data augmentation-pushing the method toward more adaptive and robust surrogate modeling.

\section{Acknowledgements}
\noindent
I.K., S.G., and L.G.B. are supported by the U.S. Department of Energy, Office of Science, Office of Advanced Scientific Computing Research, grant under Award Number DE-SC0024162. I.S. and L.G.B were supported by the Army Research Laboratory under Cooperative Agreement Number W911NF-12-2-0023, W911NF-12-2-0022, and W911NF-22-2-0014. The computational experiments were carried out at the Advanced Research Computing at Hopkins (ARCH) core facility  (rockfish.jhu.edu) hosted by Johns Hopkins University, which is supported by the National Science Foundation (NSF) grant number OAC 1920103. The views and conclusions contained in this document are those of the authors and should not be interpreted as representing the official policies, either expressed or implied, of the Army Research Laboratory or the U.S. Government. The U.S. Government is authorized to reproduce and distribute reprints for Government purposes, notwithstanding any copyright notation herein.

\bibliographystyle{elsarticle-num-names} 
\bibliography{references}

\end{document}